\pdfoutput=1

\documentclass[runningheads]{llncs}

\usepackage[subtle, wordspacing=normal, tracking=normal, bibnotes, charwidths=normal, leading, indent, lists, paragraphs=normal, mathspacing=normal, bibliography=normal]{savetrees}

\usepackage{booktabs} 
\usepackage{mdframed} 
\usepackage{paralist}
\usepackage{graphicx} 
\usepackage{enumitem}
\usepackage{balance}
\usepackage{xcolor}
\usepackage{bbold}
\usepackage{multirow}
\usepackage{tabularx}
\usepackage{amsmath,amsfonts,amssymb}
\usepackage{todonotes}
\usepackage{setspace}
\usepackage{fancyvrb}
\usepackage{times}
\usepackage[skip=0pt,bf]{caption}

\usepackage{color,soul}
\definecolor{lightblue}{rgb}{.80,.95,1}
\sethlcolor{lightblue}

\usepackage{hyperref}
\usepackage{cleveref}
\usepackage[square,comma,numbers,sort&compress,sectionbib]{natbib} 

\newcommand{\negskip}{\vspace*{-\baselineskip}}
\newcommand{\halfnegskip}{\vspace*{-.5\baselineskip}}

\begin{document}

\title{
Measuring Semantic Coherence of a Conversation
}

\author{Svitlana Vakulenko\inst{1}, Maarten de Rijke\inst{2}, Michael Cochez\inst{3,4,5}\and Vadim Savenkov\inst{1} \and Axel Polleres\inst{1,6}}

\institute{Vienna University of Economics and Business, Vienna, Austria\\
\email{\{svitlana.vakulenko, vadim.savenkov, axel.polleres\}@wu.ac.at}\\ 
\and
University of Amsterdam, Amsterdam, The Netherlands\\
\email{derijke@uva.nl}
\and
Fraunhofer FIT, 53754 Sankt Augustin, Germany
\and
Informatik 5, RWTH University Aachen, Germany
\and
Faculty of Information Technology, University of Jyvaskyla, Finland\\
\email{michael.cochez@fit.fraunhofer.de}
\and
Complexity Science Hub Vienna, Austria, and
Stanford University, CA, USA
}

\authorrunning{S.\ Vakulenko et al.}

\maketitle

\begin{abstract}

Conversational systems have become increasingly popular as a way for humans to interact with computers. 
To be able to provide intelligent responses, conversational systems must correctly model the structure and semantics of a conversation. 
We introduce the task of measuring semantic (in)coherence in a conversation with respect to background knowledge, which relies on the identification of semantic relations between concepts introduced during a conversation. 
We propose and evaluate 
graph-based and machine learning-based approaches for measuring semantic coherence using knowledge graphs, their vector space embeddings and word embedding models, as sources of background knowledge. 
We demonstrate 
how these approaches are able to uncover different coherence patterns in conversations on the Ubuntu Dialogue Corpus.

\end{abstract}


\section{Introduction}

Conversational interfaces are seeing a rapid growth in interest.
Conversational systems need to be able to model the structure and semantics of a human conversation in order to provide intelligent responses. 
The requirement conversations be coherent is meant to improve the probability distribution over possible dialogue states and candidate responses.

A conversation is an information exchange between two or more participants.\footnote{We use the terms ``dialog'' and ``conversation'' interchangeably, while ``dialog'' refers specifically to a two-party conversation.} 
An essential property of a conversation is its \textit{coherence}; \citet{de1981textlinguistics}
describe it as a ``continuity of senses.''
Coherence constitutes the outcome of a cognitive process, and is, therefore, an inherently subjective measure.
It is always relative to the background knowledge of participants in the conversation and depends on their interpretation of utterances.
Coherence reflects the ability of an observer to perceive meaningful relations between the concepts and to be critical of the new relations being introduced. 
Meaning emerges through the interaction of the knowledge presented in the conversation with the observer's stored knowledge of the world~\cite{petofi1974semantics}. 
In other words, a conversation has to be assigned an interpretation, which depends on the knowledge available to the agent.

In this paper we focus on analyzing semantic relations that hold within dialogues, i.e., relations that hold between the concepts (entities) mentioned in the course of the same dialogue.
We call this type of relation \emph{semantic coherence}. 
We focus on semantic relations but ignore other linguistic signals that make a text coherent from a grammatical point of view. 
A classic example to illustrate the difference is due to~\citet{reason:Chomsky57a}: ``Colorless green ideas sleep furiously'' -- a syntactically well formed English sentence that is semantically incoherent.

Our hypothesis is that, apart from word embeddings, recognizing concepts in the text of a conversation and determining their semantic closeness in a background knowledge graph can be used as a measure for coherence.
To this end, we propose and evaluate several approaches to measure semantic coherence in dialogues using different sources of background knowledge: both text corpora and knowledge graphs.
The contributions that we make in this paper are threefold: 
(1)~we introduce a dialogue graph representation, which captures relations within the dialogue corpus by linking them through the semantic relations available from the background knowledge;
(2)~we formulate the semantic coherence measuring task as a binary classification task, discriminating between real dialogues and generated adversary samples,\footnote{As there is no standard corpus available for this task, we test against 5 ways to generate artificial negative samples.} and 
(3)~we investigate the performance of state-of-the-art and novel algorithms on this task: top-$k$ shortest path induced subgraphs and convolutional neural networks trained using vector embeddings.

The main challenge in applying structural knowledge to natural language understanding becomes apparent when we do not just try to differentiate between genuine conversations and completely random ones, but create adversarial examples as conversations that have similar characteristics compared to the positive examples from the dataset.
Then, the results achieved using word embeddings are usually best and suggest that knowledge graph (KG) embeddings would potentially be an efficient way to harness the structure of entity relations.
However, KG embedding-based models rely on entity linking being correct and cannot easily recover from errors made at the entity linking stage compared to other graph-based approaches that we use in our experiments.

\section{Related Work}
\label{section:related}
Several lines of research are relevant to our work: discourse analysis, dialogue systems and knowledge graphs.

\subsection{Discourse analysis}

Previous work on discourse analysis demonstrates good results in recognizing discourse structure based on lexical cohesion for specific tasks such as topic segmentation in multi-party conversations~\cite{DBLP:conf/acl/GalleyMFJ03}. 
Term frequency distribution on its own already provides a strong signal for topic drift.
A more sophisticated approach to assess text coherence is based on the entity grid representation~\cite{DBLP:journals/coling/BarzilayL08}, which represents a text as a matrix that captures occurrences of entities (columns) across sentences (rows) and indicates the role entity plays in the sentence (subject, object, or other). 
This approach relies on a syntactic (dependency) parser to annotate the entity roles and is, therefore, also targeted at measuring lexical cohesion rather than semantic relations between concepts.
The de facto standard testbed for discourse coherence models is the information (sentence) ordering task~\citep{DBLP:conf/acl/Lapata03}; 
it was recently extended to a convolutional neural network-based model for coherence scoring~\cite{DBLP:conf/acl/NguyenJ17}. 
The best results to date were demonstrated by incorporating a fraction of semantic information from an external knowledge source (entity types classification) into the original entity grid model~\citep{DBLP:conf/acl/ElsnerC11a}. 
\citet{DBLP:conf/cikm/CuiLZZ17} push the state-of-the-art on the sentence ordering task by incorporating word embeddings at the input layer of a convolutional neural network instead of the entity grid.

In summary, background knowledge has been found to be able to provide a strong signal for measuring coherence in discourse.

\subsection{Dialogue systems}

In contrast to previous research focused on measuring coherence in a monologue, we consider the task of evaluating coherence in a written dialogue setting by analyzing the largest multi-turn dialogue corpus available to date, the Ubuntu Dialogue Corpus~\cite{DBLP:conf/sigdial/LowePSP15}.

Research in dialogue systems focuses on developing models able to generate or select from candidate utterances, based on previous interactions.
\citet{DBLP:journals/dad/LowePSCLP17} evaluated several baseline models on the Ubuntu Dialogue Corpus for the next utterance classification task.
Their error analysis suggests that the models can benefit from an external knowledge of the Ubuntu domain, which could provide the missing semantic links between the concepts mentioned in the course of the conversation.
This work motivated us to consider evaluating whether relations accumulated in large knowledge graphs could provide missing semantics to make sense of a conversation.

\subsection{Knowledge graphs}

Knowledge graphs (KGs) were successfully applied for disambiguating natural language text in a variety of tasks, such as information retrieval~\cite{citeulike:14281342,DBLP:conf/sigir/HasibiBGZ17} and textual entailment~\cite{silva2018recognizing}. 
They serve an important role by providing additional relations that help to bridge the lexical gap and gain a more complete understanding of the context in comparison with shallow approaches based on lexical features alone.
There was also a recent surge in development of question answering interfaces to KGs~\cite{DBLP:conf/esws/UsbeckNHKRN17,DBLP:conf/www/LukovnikovFLA17,ramngongausbeck2018}.

Our work is orthogonal to these lines of work, as it seeks to discover the potential and limitations of KGs to support natural language \emph{understanding} beyond single search queries or factoid question answering towards a holistic interactive experience, which recognizes and supports the natural (coherent) flow of a conversation.



\section{Measuring Semantic Coherence}
\label{section:approach}

In this section, we describe several approaches to modeling a conversation and measuring its coherence.
We use dialogues, i.e., a two-party conversation to illustrate our approaches.
Our approaches could also be applied to multi-party conversations.

We propose to measure dialogue coherence with a numeric score that indicates more coherent parts of a conversation and provides a signal for topic drift.
Our approach is based on the assumption that naturally occurring human dialogues, on average, exhibit more coherence than their random permutations.


\subsection{Dialogue graph}


We model a dialogue as a graph $D$, which contains 4 types of nodes ${P, U, W, C}$ and edges $E$ between them.
$P$ refers to the set of conversation participants, $U$ -- the set of utterances, $W$ -- the set of words and $C$ -- the set of concepts. 

The words $w$ in a conversation are grouped into utterances 
$\forall w \in W, \exists (u, w) \in E $ such that $u \in U$,\footnote{\label{fn:order}For simplicity, we ignore the role of word order; it can be re-constructed from the order within the conversation $T$ if needed, see below.} 
which belong to one of the conversation participants $ \forall u \in U \exists(p, u) \in E$ such that $p \in P $. 
Every utterance can belong to only one of the participants, while the same words can be re-used in different utterances by the same or different participants.
Words may refer to concepts from the background knowledge $(w, c) \in E$, where $w \in W, c \in C$. 
Several words may refer to a single concept, while the same concept may be represented by different sets of words.
The sequence in which words appear in a conversation is given by the consecutive set of edges 
$T = \{(w_1, w_2), (w_2, w_3), 
\ldots \} $ such that  $T \subset E$, 
indicating the dialogue flow.

The first three types of nodes $P$, $U$, and $W$ together with their relations are available from the dialogue transcript itself, while the set of concepts $C$ and relations between them constitute the semantic representation (meaning) of a dialogue.
The meaning is not directly observable, but is constructed by an observer (one of the dialogue participants or a third party) based on the available background knowledge.
The background knowledge supplies additional links, which we refer to as \textit{semantic relations}. They link words to concepts they may refer to: $(w, c)$ (see footnote~\ref{fn:order}) and different concepts to each other $(c_i, c_j)$. These external relations provide the missing links between words, which explain and justify their co-occurence.
The absence of such links gives an important signal to the observer, and may indicate a topic switch or discourse incoherence.
However, some of the valid links may also be missing from the background knowledge.

\begin{figure}[t]
\centering
\includegraphics[width=\textwidth]{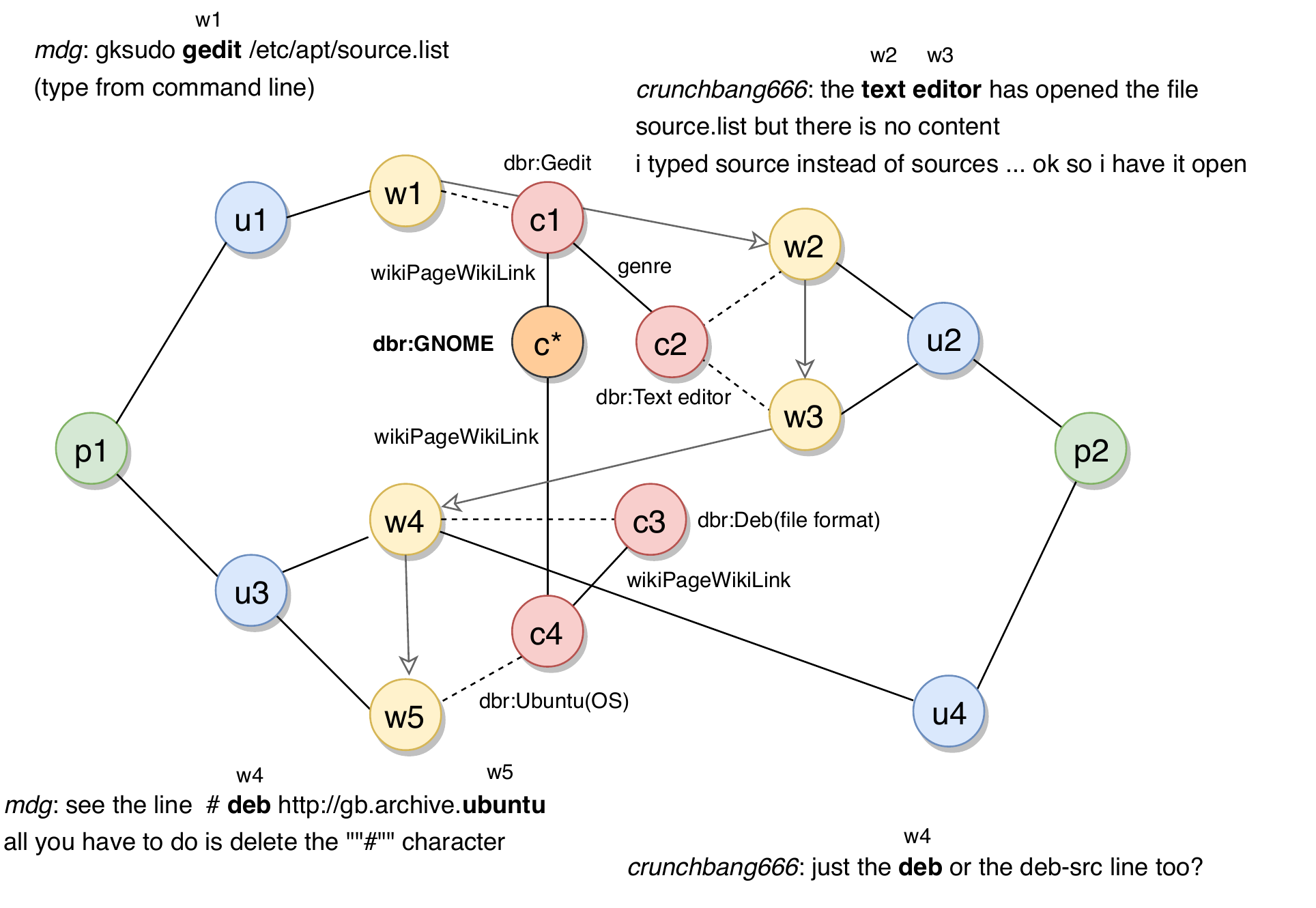}
\caption{Dialogue graph example along with the annotated dialog. We focus specifically on the layer of concepts in the middle $[c_1, \ldots, c_4]$  attempting to bridge the semantic gap in the lexicon of a conversation using available knowledge models: word embeddings and a knowledge graph.}
\label{fig:graph}
\end{figure}
An example dialogue graph is illustrated in Fig.~\ref{fig:graph}.
The dialogue consists of four utterances represented by  nodes $u1$--$u4$. In the graph we also illustrate a subgraph extracted from the background knowledge, which links the concepts $c1$ \texttt{dbr:Gedit} and $c4$ \texttt{dbr:Ubuntu(OS)} to the concept $c*$ \texttt{dbr:GNOME}, which was not mentioned in the conversation explicitly. This link represents semantic relation between the dialogue turns: ($u1, u2$) and ($u3, u4$), indicating semantic coherence in the dialogue flow. In this example, the semantic relation extracted from the background knowledge corresponds to the shortest path of length 2, i.e., the distance between the concepts mentioned in the dialogue was two relations introducing one external concept from the background knowledge. $c*$ can consist of more than one entity, but encompass a whole subgraph summarizing various relations, which hold between entities, and are represented via alternative paths between them in a knowledge graph. In the next section, we describe our approach to empirically learn semantic relations that are characteristic for human dialogues, using different sources of background knowledge and different knowledge representation models.



\subsection{Semantic relations}

We collect semantic relations between concepts referenced in a dialogue from our background knowledge.
We consider two common sources of background knowledge: (1)~unstructured data: word co-occurrence statistics from text corpora; (2)~semi-structur\-ed data: entities and relations from knowledge graphs.
In order to be able to use a KG as a source of background knowledge we need to perform an entity linking step, which maps words to semantic concepts $(w, c)$, where concepts refer to entities stored in KG.
We consider two  approaches to retrieve relations between the entities mentioned in a dialog, namely vector space embeddings and subgraph induction via the top-k shortest paths algorithm.

\negskip
\subsubsection{Embeddings.}


Embeddings are generated using the distributional hypothesis by representing an item via its context, i.e., its position and relations it holds with respect to other items.
Embeddings are multi-dimensional vectors (of a fixed size), which encode the distributional information of an item (a word in the a or a node in a graph), i.e., its position and relations to other items in the same space.
This is achieved by computing vector representations towards an optimality criteria defined with a certain output function, which depends on the embedding vectors being trained. Thus, embeddings efficiently encode (compress) an original sparse representation of the relations (e.g., an adjacency matrix) for each of the items. It provides an easy and fast way to access this information (relationship structure). Following this approach, every concept $c_i$ in our dialogue graph~\ref{fig:graph} is assigned to an n-dimensional vector, which encodes its location in the semantic space, and loses all the edges, which explicitly specified its relations to other concepts in the space.

We consider two types of embeddings to represent concepts mentioned in a dialog, one for each of our background knowledge sources: word embeddings trained on a text corpus, and entity embeddings trained on a KG. For word embeddings, we use word2vec~\cite{DBLP:conf/nips/MikolovSCCD13}, in particular the skip-gram variant, which aims to create embeddings such that they are useful for predicting words which are in the neighborhood of a given word.
GloVe~\cite{pennington2014glove} is a word embedding method, with the explicit goal of embedding analogies between entities.
This method does not work directly on the text corpus, but rather on co-occurrence counts which are derived from the original corpus.

For graph embeddings, we use two methods that can be scaled to large graphs, such as DBpedia and Wikidata: biased RDF2Vec~\cite{DBLP:conf/wims/CochezRPP17} (using random walks) and Global RDF Vector Space Embeddings~\cite{DBLP:conf/semweb/CochezRPP17};
we refer to the latter ones as \textit{KGlove} embeddings.
RDF2Vec is based on word2vec. It works by first generating random walks on the graph, where the edges have received weights which influence the probability of following these edges.
During the walk, a sentence is generated consisting of the identifiers occurring on the nodes and edges traversed.
For each entity in the graph, many walks are performed and hence a large text is generated. This text is then used for training word2vec.
KGlove is based on GloVe, but instead of counting the co-occurrence counts from text, they are computed from the graph using personalized PageRank scores starting from each node or entity in the graph.
These counts (i.e., probabilities) are then used as the input to an optimization problem that aims to encode analogies by creating embedding vectors corresponding to the co-occurrences.

\negskip
\subsubsection{Subgraph induction.}

An embedding (usually) carries a single representation for an item (word or entity), which is designed to capture all relations the item has regardless of the task or the context in which the item occurs.
For example, an embedding representation may neglect some of the infrequent relations, which can become more relevant than others depending on the situation (context).
In order to contrast the embedding-based approach, we also implement a more traditional graph-based approach to represent entity relations in a KG.
Given a sequence of entities, as they appear in a dialog, i.e., $[c_1, c_2 \ldots c_n]$, we extract relations, as top-k shortest paths, between every entity $c_i$ and all the entities that were mentioned in the same dialogue before $c_i$, i.e., $(c_1, c_i), (c_2, c_i), \ldots, (c_{i-1}, c_i)$. 

For the top-k shortest path computation, we apply an approach based on bidirectional breadth-first search~\cite{SavenkovMUP17} using the space-efficient binary {Header, Dictionary, Triples} (HDT) encoding~\cite{fernandez2013} of the KG.
This approach maps entities discussed in the dialogue to KG concepts, and then interprets paths  between concepts in the KG as semantic relations between the respective entities.
Many such relations are never mentioned in the conversation and only become explicit through the path enumeration over the KG. By increasing the number of desired shortest paths $k$ and the maximum path length $\ell$, one can discover more relations, including those that might be omitted or obscured in the entity embedding representation in the case of a random walk or frequency-based embedding algorithms.
An obvious downside of this increase in recall is reduced efficiency.

\subsection{Dialogue classification}
We measure semantic coherence by casting the task into a classification problem. The score produced by the classifier 
corresponds to our measure of semantic coherence. 

Since human dialogues are expected to exhibit a certain degree of incoherence due to topic drift and since relations are missing from our background knowledge, we cannot assume every concept in our dialogue dataset to be coherent with respect to the other concepts in the same dialog.
However, it is reasonable to assume that on average a reasonably large set of concepts extracted from a human dialogue exhibits a higher degree of coherence than a randomly generated one.
We build upon this assumption and cast the task of measuring semantic coherence as a binary classification task, in which real dialogues have to be distinguished from corrupted (incoherent) dialogues. 
We consider positive and negative examples for whole conversations, represented as a sequence of words or entities, which constitute the input for the binary classifier. 
Effectively, these examples provide a supervision signal for measuring and aggregating distances between words/concepts by learning the weights for the neural network classifier.

\negskip
\subsubsection{Negative sampling.}
\label{section:sampling}

To produce negative (adversarial) examples for the binary classification task we propose five sampling strategies:

\begin{itemize}[nosep]
\item RUf: {Random uniform}. For every positive example we choose a sequence of entities (or words for training on word embeddings) of the same size from the vocabulary uniformly at random; so, we double the size of the dataset effectively by supplementing it with completely randomly generated (i.e., presumably incoherent) counterexamples.

\item SqD: {Sequence disorder}. Randomly permute the original sequence, which is similar in spirit to the sentence ordering task for evaluating discourse coherence~\cite{DBLP:conf/acl/Lapata03}. The key difference is that we rearrange the order of words (entities), which may also occur within the same sentence (utterance), rather than permuting whole sentences. 

\item VoD: {Vocabulary distribution}. For every positive example choose a sequence of entities of the same length from the vocabulary using the same frequency distribution as in the original corpus; so, VoD is very similar to RuF, but tries to emulate ``structure'' to some extent by choosing similar term frequencies.

\item VSp: {Vertical split}. Create a negative example by permuting two positive examples replacing utterances of one of the conversation participants with utterances of a participant from a different conversation.

\item HSp: {Horizontal split}. Create a negative example by permuting two positive examples merging the first half of one conversation with the second half of a different conversation.
\end{itemize}

\negskip
\subsubsection{Convolutional neural network.}

To solve the binary classification task we train a classifier using a convolutional neural network architecture, which is applied to sequences of words and entities to distinguish irregular semantic drift, which was deliberately injected into conversations, from smooth drift which occur within real conversations.

It is a standard architecture previously employed for a variety of natural language tasks, such as text classification~\cite{DBLP:journals/corr/Kim14f}. The network consists of (1) an input layer, which appends the pre-trained embeddings for each of the word (entity) from the dialogue sequence; (2) a convolutional layer, which consist of filters (arrays of trainable weights) sliding over and learning predictive local patterns in the previous layer of the input embeddings; (3) a max pooling layer, which aggregates the features learned by the neighboring filters; (4) the hidden layer, a fully connected layer, which allows combining features from all the dimensions with a non-linear function; and (5) the output layer is a fully connected layer, which aggregates the scores to make the final prediction.
See also Section~\ref{sec:implementation} for details.








\section{Evaluation Setup}
\label{section:evaluation}

The source code of our implementation and evaluation procedures is publicly accessible.\footnote{\url{https://github.com/vendi12/semantic\_coherence}} We also release our dataset used in the evaluation, which contains dialogue annotations with DBpedia entities and shortest paths, for reproducibility and further references.\footnote{\url{https://github.com/vendi12/semantic_coherence/tree/master/data}}

\subsection{Dataset}

\halfnegskip
\subsubsection{Dialogues.}
Our experiments were performed on a sample of dialogues from the Ubuntu Dialogue Corpus\footnote{\url{https://github.com/rkadlec/ubuntu-ranking-dataset-creator}}~\cite{DBLP:conf/sigdial/LowePSP15}, which contains 1,852,869 dialogues in total, with one dialogue per file in TSV format, and is the largest conversational dataset to date. 
There are multiple challenges related to using this corpus, however. 
The dialogues were automatically extracted from a public chat using several heuristics selecting two user handles and segmenting based on the timestamps. 
The dialogues cannot be considered as perfectly coherent since some of the related utterances are missing from the dialogues; there can be several different topics discussed within the same conversation and the asynchronous nature of on-line communication often results in semantic mismatch in the dialogue sequence. 
While we cannot guarantee local coherence of the real dialogues, we expect them to be on average more coherent, when comparing to the dialogues randomly generated by sampling entities (words) from the vocabulary or merging entities (words) from different dialogues, which we refer to as negative samples, or adversaries, in our binary classification task.

We proceed by annotating a sample of 291,848 dialogues from the Ubuntu Dialogue Corpus with the DBpedia entities using the DBpedia Spotlight public web service\footnote{\url{http://model.dbpedia-spotlight.org/en/annotate}}~\cite{isem2013daiber}. 
The input to the entity linking API is the text for each utterance in a conversation. 
Next, we considered only the dialogues where both participants contribute at least 3 new entities each, i.e., every dialogue in our dataset contains minimum 6 entities shared between the dialogue partners. 
The threshold for entities per conversation was chosen to ensure there is enough semantic information for measuring coherence. 
This way, we end up with a sample of 45,510 dialogues, which we regard as true positive examples of coherent dialogue. It contains 17,802 distinct entities and 21,832 distinct words that refer to these. 
The maximum size of a dialogue in this dataset is 115 entities or 128 words referring to them. We shuffled the dialogues and selected 5,000 dialogues for our test set. 
While this procedure means we cannot test our approach on short conversations, with fewer entities, we consider 45K dialogues to be a representative dataset for evaluating our approach.

The negative samples for both training and test set were generated using five different sampling strategies described in Section~\ref{section:sampling}. Each development set consists of 81,020 samples (50\% positive and 50\% negative). 
We further split it into a training and validation set: 64,816 and 16,204 (20\%) samples, respectively. 
Our test set comprises the remaining 5,000 positive examples, and 5,000 generated negative samples.

\negskip
\subsubsection{Knowledge models.}

We compared the performance on our task across two types of embeddings models trained on two different knowledge source types: GloVe~\cite{pennington2014glove} and Word2Vec~\cite{DBLP:conf/nips/MikolovSCCD13} for the word embeddings, and biased RDF2vec~\cite{DBLP:conf/wims/CochezRPP17} and KGloVe~\cite{DBLP:conf/semweb/CochezRPP17} for the knowledge graph entity embeddings.

We utilise two publicly available word embedding models: GloVe embeddings pre-trained on the Common Crawl corpus (2.2M words, 300 dimensions)\footnote{\url{https://nlp.stanford.edu/projects/glove/}} and Word2Vec model trained on the Google News corpus (3M words, 300 dimensions).\footnote{\url{https://code.google.com/archive/p/word2vec/}} 1,578 words from our dialogues (7\%) were not found in the GloVe embeddings dataset and received a zero vector in our embedding layer. Thus, GloVe embeddings cover 20,254 words from our vocabulary (93\%). Word2Vec embeddings cover only 73\% of our vocabulary.

For RDF2Vec and KGloVe we used publicly available pre-trained global embeddings of the DBpedia entities (see~\cite{DBLP:conf/wims/CochezRPP17} and~\cite{DBLP:conf/semweb/CochezRPP17}, respectively). 
For KGlove we used all different embeddings, while for RDF2Vec we experimented with the embeddings that gave the best performance in \cite{DBLP:conf/wims/CochezRPP17}.
KGlove embeddings cover 17,258 entities from our vocabulary (97\%), while
Rdf2Vec provides 62--77\% due to different importance sampling strategies of the embedding approaches.

The shortest paths used were extracted from dumps of DBpedia (April 2016, 1.04 billion triples) and Wikidata (March 2017, 2.26 billion triples).\footnote{\url{http://www.rdfhdt.org/datasets/}}

\subsection{Implementation}
\label{sec:implementation}
Our neural network model is implemented using the Keras library with a TensorFlow backend. 
The one-dimensional (temporal) convolutional layer contains 250 filters of size 3 and stride (step) 1. The max pooling layer is global, the hidden layer is set to 250 dimensions.
There are two activation layers with rectified linear unit (ReLU) after the convolutional and the hidden layers to capture also non-linear dependencies between input and output, and two dropout layers with rate 0.2 after the embeddings and hidden layers to avoid overfitting.
The last ReLU activation is projected onto a single-unit output layer with a sigmoid activation function to obtain a coherence score on the interval between 0 and 1.

The network is trained using the Adam optimizer with the default parameters~\cite{DBLP:journals/corr/KingmaB14} to minimize the binary cross-entropy loss between the predicted and correct value.
All models were trained for 10 epochs in batches of 128 samples and early stopping after 5 epochs if there is no improvement in accuracy on the validation set.

To compute the shortest paths we merged the dumps of DBpedia and Wikidata into a single 36GB binary file in  HDT format~\cite{fernandez2013} (DBpedia+Wikidata HDT), with an additional 21GB index on the subject and the object components of triples.
We set the parameters of the algorithm in our experimental evaluation as follows: $k$ for the number of shortest paths to be retrieved from the graph to 5, the maximum length $\ell$ of a path to 9 edges (relations) and a timeout terminating the query after 2 seconds to cope with the scalability issues of the algorithm.
Our top-k shortest paths algorithm implementation is available via a SPARQL endpoint\footnote{\url{http://wikidata.communidata.at}} using the syntax shown in Fig.~\ref{fig:k-shortest}.

\begin{figure}[ht]
\begin{Verbatim}[fontsize=\small]
 PREFIX ppf: <java:at.ac.wu.arqext.path.>
 PREFIX dbr: <http://dbpedia.org/resource/>                
 SELECT * WHERE { 
 ?X ppf:topk ("--source" dbr:Directory_service 
                         dbr:Gnome dbr:GNOME 
                         dbr:Desktop_environment 
              "--target" dbr:Desktop_computer 
              "--k" 5 "--maxlength" 9 "--timeout" 2000) }\end{Verbatim}
\caption{\label{fig:k-shortest} k-shortest path query (cf.~\cite{SavenkovMUP17} to extract relevant connections between entities from the knowledge graph}
\end{figure}
 The function \texttt{at.ac.wu.arqext.path.topk} is a user defined extension available as a Jena ARQ extension.\footnote{\url{https://bitbucket.org/vadim\_savenkov/topk-pfn}}

\section{Evaluation Results}
\label{section:results}

Table~\ref{tbl:context} reports the most common entities and relations, which while not being mentioned in the course of a dialogue, were on the shortest paths (in the KG) between other entities that were explicitly mentioned in the dialogue, i.e., which constitute an implicit dialogue context. 
While Dbpedia Spotlight dereferenced ``Ubuntu'' mentions to the concept related to philosophy rather than to the popular software distribution, the graph-based approach succeeds in recovering the correct meaning of the word by extracting this concept from the shortest paths that lie between the other entities mentioned in dialogues. Almost all relations obtained from the KG correspond to the links between the corresponding Wikipedia web pages (wikiPageWikiLink).

\begin{table}[t]
\centering
\caption{The top 5 most common entities and relations in the Ubuntu Dialogue dataset: mentioned entities -- from linking dialogue utterances to DBpedia entities via Dbpedia Spotlight Web service; context entities and relations -- from the shortest paths between the mentioned entities in DBpedia.}
\label{tbl:context}
\begin{tabular}{clrlrlr}
\toprule
Top & \multicolumn{2}{c}{Mentioned entities} & \multicolumn{2}{c}{Context entities} & \multicolumn{2}{c}{Relations}     \\
\cmidrule(r){2-3}
\cmidrule(r){4-5}
\cmidrule{6-7}
\#  & Label                     & Count      & Label                        & Count & Label                     & Count \\
\midrule
1                       & Ubuntu(philosophy)      & 1605                      & Ubuntu(OS) & 1058                      & wikiPageWikiLink & 51014                     \\
2                       & Sudo                      & 708                       & Linux                       & 725                       & gold/hypernym             & 319                       \\
3                       & Booting                   & 676                       & Microsoft\_Windows          & 208                       & ontology/genre            & 178                       \\
4                       & APT(Debian)             & 405                       & FreeBSD                     & 175                       & operatingSystem  & 140                       \\
5                       & Live\_CD                  & 314                       & Smartphone                  & 171                       & rdf-schema\#seeAlso       & 116        \\              
\bottomrule
\end{tabular}
\end{table}

\subsection{Semantic distance}

\begin{figure}[!t]
\centering
\includegraphics[width=0.76\textwidth]{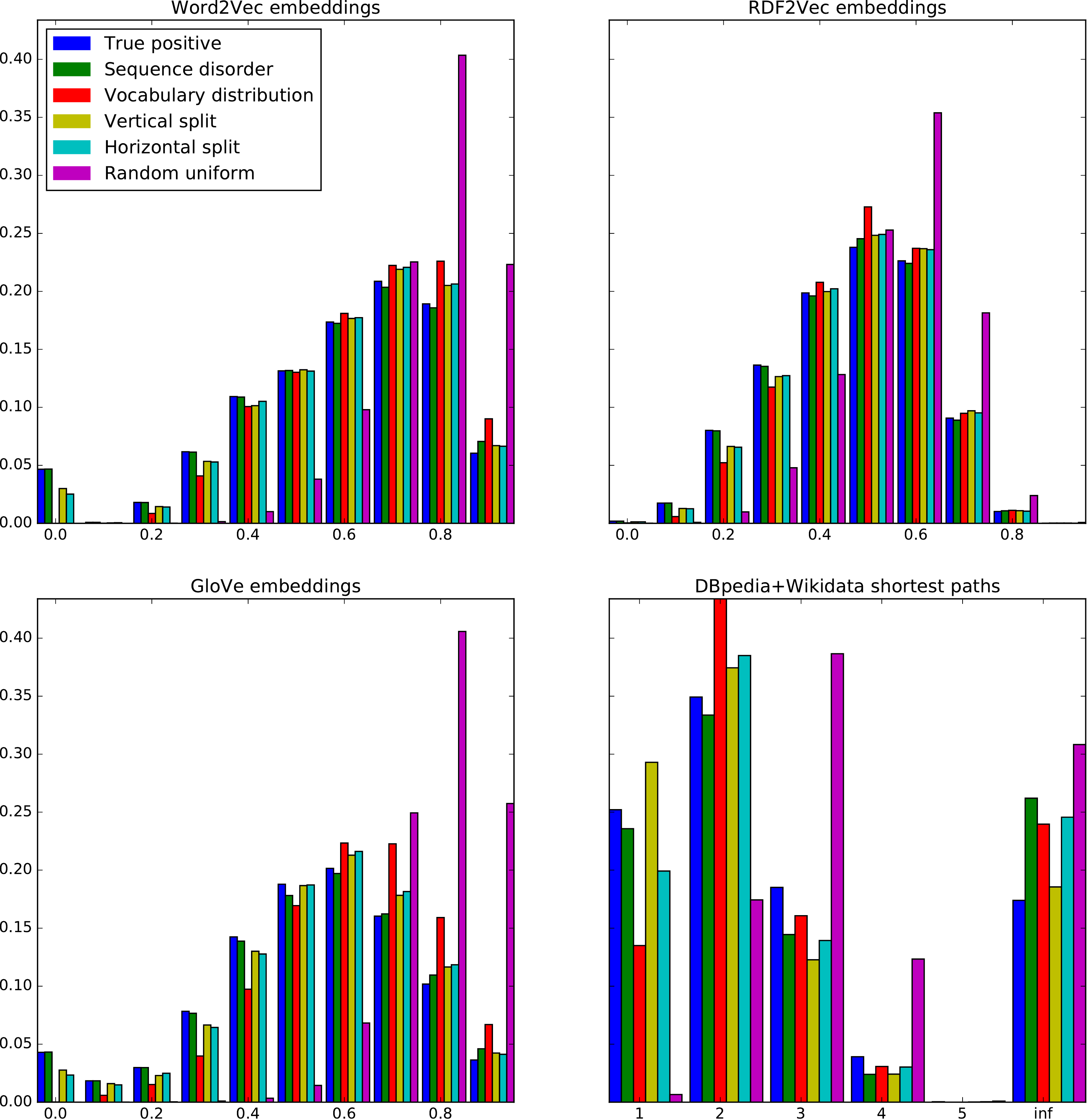}
\caption{Distribution of cosine distances for different data splits using Word2Vec and GloVe word embeddings (left), and RDF2Vec KG embeddings (top right), compared with the distribution of shortest path lengths in DBpedia+Wikidata KG (bottom right). Words in real dialogues (True positive) are more related than frequent domain words (Vocabulary distribution), and much more than a random sample (Random uniform).}
\label{fig:sem_dist_distrs}
\end{figure}

The length of the shortest path (number of edges, i.e., relations on the path) is a standard measure used to estimate semantic (dis)similarity between entities in a knowledge graph~\cite{DBLP:journals/corr/abs-1203-1889}. We observe how it correlates with a standard measure to estimate similarity between vectors in a vector space, \textit{cosine distance}, defined as: 
$1 - \cos(x, y) = 1 -\frac {x y^\intercal}{|| x||  || y||}$ 
Fig.~\ref{fig:sem_dist_distrs} showcases different perspectives on semantic similarity (coherence) between the entities in real and generated dialogues as observed in different semantic spaces (w.r.t. the knowledge models), alignments and differences between them. The barplots reflect the distributions of the semantic distances between entities in dialogues. The semantic distances are measured using cosine distances between vectors in the vector space for word (Word2Vec and GloVe) and KG (RDF2Vec) embeddings, and in terms of the shortest path lengths in the DBpedia+Wikidata KG. We observe that the real dialogues (True positive) tend to have smaller distances between entities: 1--2 hops or at most 0.3 cosine distance, while randomly generated sequences are skewed further off. Embeddings produce much more fine-grained (continuous) representation of semantic distances in comparison with the shortest path length metric. Distributions produced by different word embeddings are very similar in shape, while the one from KG embeddings is steeper and skewed more to the center, there are only a few entities further than 0.7, while this is the top for the random distances in word embeddings.

We also discover the bottleneck of our shortest path algorithm at length 5. Since the set of relevant entities for which the paths are computed grows proportionally to the dialogue length, depending on the degree of the node the number of expanded nodes quickly reaches the limit on the memory size.  In our case, the algorithm retrieved the paths of length at most 5 due to the 2-second timeout, while the parameter for the maximum length of the path $\ell$ was set to 9.

\subsection{Classification results}
Our evaluation results from training a neural network on the task of measuring (in)cohe\-rence in dialogues are listed in Table~\ref{tbl:results}. It summarizes the outcomes of models trained on different embeddings using different types of adversarial samples (negative sampling strategies are described in Section~\ref{section:sampling}). For the KG embeddings, we report only the approaches that performed best across different test splits.\footnote{The full result table is available on-line: \url{https://github.com/vendi12/semantic_coherence/blob/master/results/results.xls}}

\begin{table}[!t]
\centering
\caption{Accuracy on the test set across different embedding and sampling approaches. 
The table shows for 7 different embedding strategies (4 types), how the embedding performs when trained with data from different generated adversarial examples.
For example, the underlined value in the table (0.92), means that GloVe word embeddings, when trained with genuine and Vertical split (VSp) adversarial examples, is able to correctly find 92\% of the Vocabulary distribution (VoD) adversarial examples in the test set.
In the same row, in the TPos column, it can be seen that 60\% of the genuine messages were correctly identified. Hence, this results in an average accuracy of 0.76.
In blue highlight, we indicate the results where the adversarial examples for training the model where of the same type as for testing the model.
In bold, we indicate the best result for each adversarial example type.
	Abbreviations: TPos -- True Positive, TNeg -- True Negative, RUf -- Random uniform, VoD -- Vocabulary distribution, SqD -- Sequence disorder, VSp -- Vertical split, HSp -- Horizontal split, Avg -- Average, PRS -- PageRank Split, PR -- PageRank, Uni -- Uniform, PrO -- Predicate Object.}
\label{tbl:results}
\begin{tabular}{p{2cm}lllllllllllll}
\toprule
 &  & \multicolumn{12}{c}{Accuracy}                                                                                      \\ 
\cmidrule{3-14}
                                  &  Data &  & \multicolumn{10}{c}{TNeg}                                           &  \\ 
    \cmidrule{4-13}
Embeddings & split & TPos & RUf  & Avg  & VoD  & Avg  & SqD  & Avg  & VSp  & Avg  & HSp  & Avg  & Avg  \\ 
    \midrule
Word2Vec                          & RUf                         & 0.99                  & 0.99 & \hl{\textbf{0.99}} & 0.02 & 0.50 & 0.02 & 0.50 & 0.01 & 0.50 & 0.01 & 0.50 & 0.60                 \\
                                  & VoD                         & 0.89                  & 0.62 & 0.75 & 0.90 & \hl{0.89} & 0.53 & 0.71 & 0.18 & 0.54 & 0.20 & 0.54 & \textbf{0.69}                 \\
                                  & SqD                         & 0.75                  & 0.65 & 0.70 & 0.88 & 0.81 & 0.81 & \hl{0.78} & 0.27 & 0.51 & 0.29 & 0.52 & 0.66                 \\
                                  & VSp                         & 0.59                  & 0.50 & 0.55 & 0.82 & 0.71 & 0.41 & 0.50 & 0.59 & \hl{0.59} & 0.61 & 0.60 & 0.59                 \\
                                  & HSp                         & 0.62                  & 0.39 & 0.50 & 0.71 & 0.66 & 0.38 & 0.50 & 0.55 & 0.58 & 0.63 & \hl{0.63} & 0.58                 \\ 
\midrule
GloVe                       & RUf                         & 0.99                  & 0.99 & \hl{\textbf{0.99}} & 0.00 & 0.50 & 0.01 & 0.50 & 0.00 & 0.50 & 0.00 & 0.50 & 0.60                 \\
                                  & VoD                         & 0.93                  & 0.38 & 0.66 & 0.93 & \hl{\textbf{0.93}} & 0.39 & 0.66 & 0.19 & 0.56 & 0.08 & 0.51 & 0.66                 \\
                                  & SqD                         & 0.76                  & 0.71 & 0.73 & 0.91 & 0.84 & 0.82 & \hl{\textbf{0.79}} & 0.16 & 0.46 & 0.15 & 0.45 & 0.66                 \\
                                  & VSp                         & 0.60                  & 0.25 & 0.42 & \underline{0.92} & 0.76 & 0.43 & 0.51 & 0.65 & \hl{0.62} & 0.66 & 0.63 & 0.59                 \\
                                  & HSp                         & 0.71                  & 0.34 & 0.52 & 0.81 & 0.76 & 0.30 & 0.50 & 0.55 & \textbf{0.63} & 0.66 & \hl{\textbf{0.68}} & 0.62                 \\ 
\midrule
rdf2vec PRS            & RUf                         & 0.98                  & 0.99 & \hl{\textbf{0.99}} & 0.02 & 0.50 & 0.02 & 0.50 & 0.02 & 0.50 & 0.01 & 0.50 & 0.60                 \\
                                  & VoD                         & 0.79                  & 0.68 & 0.73 & 0.83 & \hl{0.81} & 0.34 & 0.57 & 0.36 & 0.57 & 0.35 & 0.57 & 0.65                 \\
                                  & SqD                         & 0.59                  & 0.48 & 0.54 & 0.72 & 0.66 & 0.67 & \hl{0.63} & 0.43 & 0.51 & 0.40 & 0.50 & 0.56                 \\
rdf2vec PR                 & HSp                         & 0.57                  & 0.59 & 0.58 & 0.72 & 0.64 & 0.43 & 0.50 & 0.59 & 0.58 & 0.67 & \hl{0.62} & 0.58                 \\ 
\midrule
KGloVe Uni        & RUf                         & 0.92                  & 0.97 & \hl{0.94} & 0.11 & 0.51 & 0.09 & 0.50 & 0.08 & 0.50 & 0.07 & 0.50 & 0.59                 \\
                                  & VoD                         & 0.54                  & 0.88 & 0.71 & 0.73 & \hl{0.64} & 0.61 & 0.58 & 0.51 & 0.52 & 0.52 & 0.53 & 0.60                 \\
                                  & SqD                         & 0.55                  & 0.62 & 0.58 & 0.64 & 0.59 & 0.63 & \hl{0.59} & 0.47 & 0.51 & 0.45 & 0.50 & 0.56                 \\
KGloVe PrO & HSp                         & 0.31                  & 0.81 & 0.56 & 0.75 & 0.53 & 0.69 & 0.50 & 0.77 & 0.54 & 0.70 & \hl{0.51} & 0.53                 \\
KGloVe PR     & HSp                         & 0.47                  & 0.69 & 0.58 & 0.61 & 0.54 & 0.54 & 0.50 & 0.57 & 0.52 & 0.65 & \hl{0.56} & 0.54     \\ 
\bottomrule
\end{tabular}
\end{table}

From the results we observe that the easiest task was to distinguish real dialogues from randomly generated sequences.
When the model was trained with randomly generate dialogues, accuracies often reach close to 100\%.
However, this same model performs poorly when used for any other type of non-genuine messages we created. 
In the best case (KGloVe Uni), still only 10\% of messages randomly sampled from the vocabulary distribution were correctly detected.
This indicates that there is a need to experiment with the other types as well.
We also observe that the models that are trained with specific adversarial examples are best in separating that type.
However, even when the model is not explicitly trained to recognize a specific type of dialogue, but instead trained on other types of adversarial examples, it is sometimes still able to classify  messages correctly. 
This happens, for example, in the case of KGloVe Uniform where the adversarial messages are sampled from the Vocabulary distribution and the model is still able to detect around 70\% of randomly generated messages.

The dialogues generated by permuting the sequence of entities (words) in the original dialogues (the sequence ordering task) were harder to distinguish (The best performing model resulted in an accuracy of 0.79). 
Finally, the hardest task was to discriminate the adversarial examples generated by merging two different dialogues together (vertical and horizontal splits).
This was expected as these dialogues have short sequences of genuine dialogue inside, making them hard to classify.

The best performance across all test settings was achieved using the word embeddings models, especially GloVe performed well. 
KG embeddings, while performing reasonably well on the easier tasks (RUf and VoD), fell short to distinguish more subtle changes in semantic coherence.
For the KG embedding weighting approaches, we noticed that the ones which performed well in earlier work, also worked better in this task.
In particular, it was noticed that the weighting biased by PageRank computed on the Wikipedia links graph results in better results in machine learning tasks.

As discussed in Section \ref{section:evaluation}, RDF2vec has fewer entity embeddings than KGloVe, when trained from the same original graph (DBpedia). KGloVe will provide an embedding, even when not much is known about a specific entity.
In case of a node that does not have any edges, KGloVe will assign a random vector to it.
In contrast, RDF2Vec will prune infrequent nodes.
Another problem that affects KG embeddings are incorrectly recognized entities.
There is no linking required for needed word embeddings since it represents different meanings of the word in a single vector.



\begin{figure}[!t]
\centering

\includegraphics[width=0.92\textwidth]{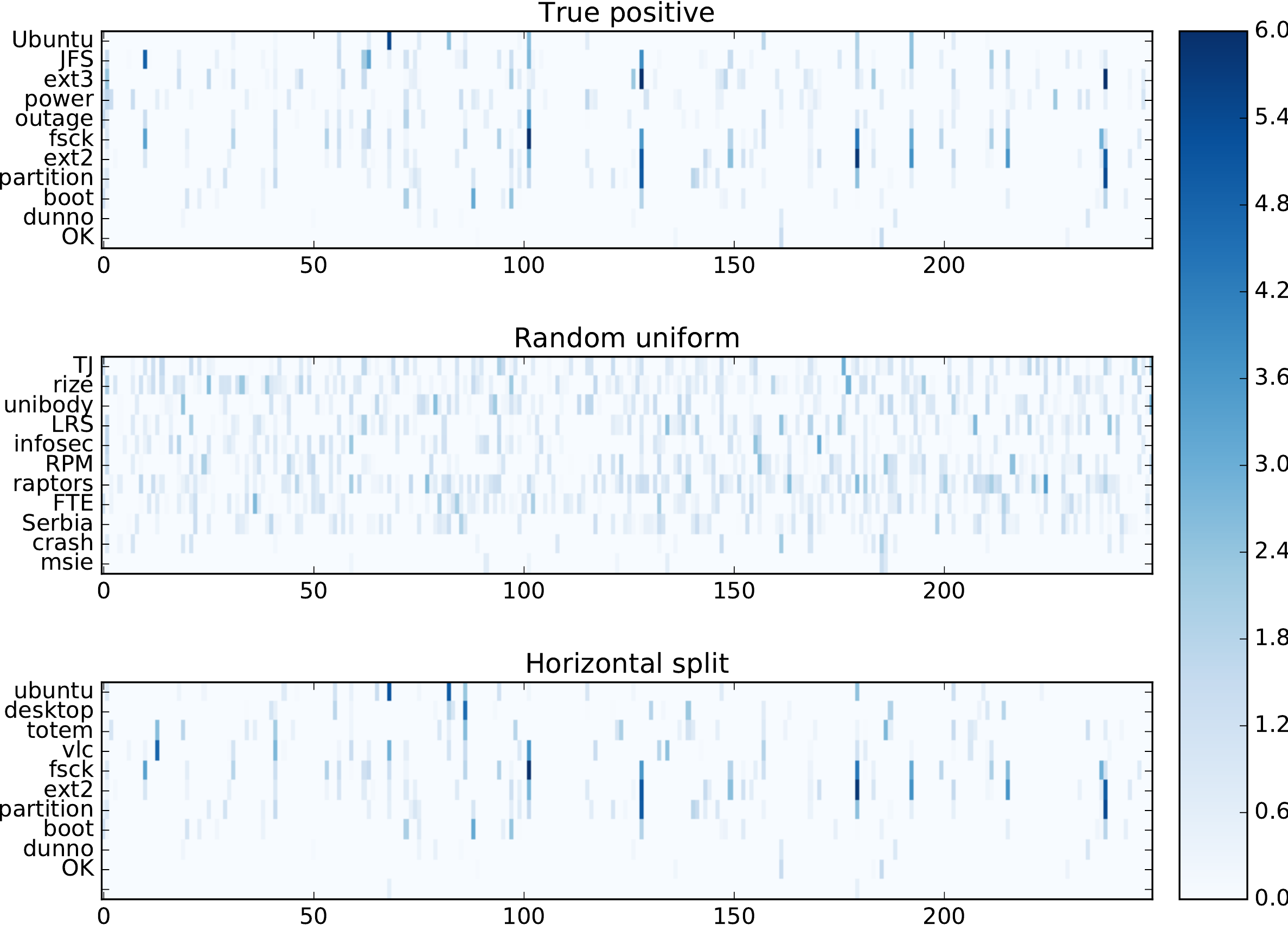}
\caption{Heatmap of the activations on the output of the word embeddings layer. Notice the vertical-bar pattern indicating a stronger semantic relation between the words in a real dialogue (top) in comparison with a random word sequence (middle). The topic drift effect can be observed when two different dialogues are concatenated (horizontal split -- bottom): the bars at the top are shifted in comparison with the bars in the second half of the conversation, comparing to the coherence patterns observed in the real dialogue (top).}

\label{fig:heatmap}
\end{figure}

Overall, we want to be able not only to tell to which degree a dialogue is (in)coherent but also to identify the regions in the dialogue where coherence was disrupted, or to partition the dialogue into coherent segments indicating the shifts between different topics. 
Visualization of the activations in the output of the convolutional layer of the Glove word embeddings-based model exhibits distinct vertical activation patterns, which can be interpreted as traces of local coherence the model is able to recognize (See Fig.~\ref{fig:heatmap}).



\section{Conclusion}
\label{section:conclusion}

We considered the task of measuring semantic coherence of a conversation, which introduces an important and challenging problem that requires operating vast amounts of heterogeneous knowledge sources to infer implicit relations between the utterances, i.e., bridging the semantic gap in understanding natural language.
We proposed and evaluated several approaches to this problem using alternative sources of background knowledge, such as structured (knowledge graph) and unstructured (text corpora) knowledge representations. 
These approaches detect semantic drift in conversations by measuring  coherence with respect to the background knowledge. 
Our models were trained for dialogs but the approach does not restrict the number of conversation participants. 
The model's performance depends to a large extent on the choice of background knowledge source, with respect to the conversation domain. 
The conversation needs to contain a sufficient number of recognized entities to signal its position within the semantic space.

Our results indicate promising directions as well as challenges in applying structural knowledge to analyse natural language. We show that the use of word embeddings in text classification is superior to some existing knowledge graph embeddings. This is an important insight, advancing research by uncovering limitations of state-of-the-art knowledge graph embeddings and indicating directions for improvements.

Knowledge graph embeddings constitute a potentially powerful method to efficiently harness entity relations for tasks that require estimates of semantic similarity. However, their use relies on the correctness of the entity linking performance. 
Errors made at this stage in the pipe-line approach do propagate into the classification results, but we noticed that they are rather consistent, which partially mitigates the problem.
Our experiments showed that graph-based approaches are more robust to errors in entity linking than knowledge graph embeddings, which is an important insight for future work: this effect can likewise be expected with other existing entity linking approaches.

More research is needed on how to make a knowledge graph embeddings-based model more robust to uncertainty in entity linking, such as end-to-end learning on graphs~\cite{wilcke2017knowledge}. Also, combining evidence from both structured (knowledge graphs) and unstructured (text) data sources has a great potential to mitigate knowledge sparsity, increase support and interpretability of semantic relations~\cite{DBLP:conf/semweb/ThomaRB17}. We provide a test bed for the semantic coherence task, which can be used to compare word- and entity-based representation approaches, and their combinations, whereupon others can build.

\begin{spacing}{1}
\smallskip\noindent\small
\textbf{Acknowledgments.}
This work is supported by the project 855407 ``Open Data for Local Communities'' (CommuniData) of the Austrian Federal Ministry of Transport, Innovation and Technology (BMVIT) under the program ``ICT of the Future.''
Svitlana Vakulenko was supported by the EU H2020 programme under the MSCA-RISE agreement 645751 ({RISE\_BPM}).
Axel Polleres was supported under the Distinguished Visiting Austrian Chair Professors program hosted by The Europe Center of Stanford University.
Maarten de Rijke was supported by 
Ahold Delhaize,
Amsterdam Data Science,
the Bloomberg Research Grant program,
the China Scholarship Council,
the Criteo Faculty Research Award program,
Elsevier,
the European Community's Seventh Framework Programme (FP7/2007-2013) under
grant agreement nr 312827 (VOX-Pol),
the Google Faculty Research Awards program,
the Microsoft Research Ph.D.\ program,
the Netherlands Institute for Sound and Vision,
the Netherlands Organisation for Scientific Research (NWO)
under pro\-ject nrs
CI-14-25, 
652.\-002.\-001, 
612.\-001.\-551, 
652.\-001.\-003, 
and
Yandex.
All content represents the opinion of the authors, which is not necessarily shared or endorsed by their respective employers and/or sponsors.
\end{spacing}

\vspace{-1ex}

\bibliographystyle{splncsnat}
\bibliography{refs} 

\end{document}